%% file: main.tex
\documentclass[letterpaper, 10 pt, conference]{ieeeconf}  

\IEEEoverridecommandlockouts                              

\usepackage{multirow}

\usepackage{graphicx}
\usepackage{booktabs}
\usepackage[table]{xcolor}

\usepackage{amsmath} 
\usepackage{amssymb}  
\usepackage{xcolor}
\usepackage{siunitx}
\usepackage{svg}
\usepackage{pdfpages}
\definecolor{rblue}{rgb}{0,0.5,1}
\definecolor{awesome}{rgb}{1.0, 0.13, 0.32}
\definecolor{hollywoodcerise}{rgb}{0.96, 0.0, 0.63}
\definecolor{lasallegreen}{rgb}{0.03, 0.47, 0.19}
\definecolor{hanpurple}{rgb}{0.32, 0.09, 0.98}
\definecolor{green(pigment)}{rgb}{0.0, 0.65, 0.31}
\usepackage{tikz}
\newcommand*\circled[1]{\tikz[baseline=(char.base)]{
\node[shape=circle,fill=gray,inner sep=0.598pt] (char) {\textcolor{white}{\small \textbf{#1}}};}}

\def\eg{\emph{e.g.}}

\makeatletter
\let\NAT@parse\undefined
\makeatother
\usepackage[pagebackref=false,breaklinks=true,colorlinks,bookmarks=false]{hyperref}
\hypersetup{colorlinks=true,
    linkcolor={red},
    citecolor={hanpurple},
    urlcolor={magenta}
}

\title{\LARGE \bf
MICA: Multi-Agent Industrial Coordination Assistant
}

\author{
Di Wen$^{1}$, Kunyu Peng$^{1,2,*}$, Junwei Zheng$^{1}$, Yufan Chen$^{1}$, Yitian Shi$^{1}$, Jiale Wei$^{1}$, Ruiping Liu$^{1}$,\\Kailun Yang$^{3}$, and Rainer Stiefelhagen$^{1}$%
\thanks{This work was supported in part by the SmartAge project sponsored by the Carl Zeiss Stiftung (P2019-01-003; 2021-2026), the University of Excellence through the ``KIT Future Fields'' project, in part by the Helmholtz Association Initiative and Networking Fund on the HoreKA@KIT partition and the state of Baden-Württemberg through bwHPC and the German Research Foundation (DFG) through grant INST 35/1597-1 FUGG. 
This work was also supported in part by the National Natural Science Foundation of China (Grant No. 62473139), in part by the Hunan Provincial Research and Development Project (Grant No. 2025QK3019), and in part by the State Key Laboratory of Autonomous Intelligent Unmanned Systems (the opening project number ZZKF2025-2-10).}%
\thanks{$^{1}$The authors are with Karlsruhe Institute of Technology, Germany.}%
\thanks{$^{2}$The author is also with INSAIT, Sofia University ``St. Kliment Ohridski'', Bulgaria.}
\thanks{$^{3}$The author is with Hunan University, China.}%
\thanks{*Corresponding author: Kunyu Peng (kunyu.peng@kit.edu).}%
}

\begin{document}

\bstctlcite{MyBSTcontrol} %

\maketitle
\thispagestyle{empty}
\pagestyle{empty}

\input{tex/Abstract}
\input{tex/Introduction}

\input{tex/RelatedWork}
\input{tex/Methodology}
\input{tex/Evaluation}

\input{tex/Discussion}

\bibliographystyle{IEEEtran}
\bibliography{IEEEabrv,refs}

\end{document}

%% file: tex/Abstract.tex
\begin{abstract}
Industrial workflows demand adaptive and trustworthy assistance that can operate under limited computing, connectivity, and strict privacy constraints. In this work, we present MICA (Multi-Agent Industrial Coordination Assistant), a perception-grounded and speech-interactive system that delivers real-time guidance for assembly, troubleshooting, part queries, and maintenance. MICA coordinates five role-specialized language agents, audited by a safety checker, to ensure accurate and compliant support. To achieve robust step understanding, we introduce Adaptive Step Fusion (ASF), which dynamically blends expert reasoning with online adaptation from natural speech feedback. Furthermore, we establish a new multi-agent coordination benchmark across representative task categories and propose evaluation metrics tailored to industrial assistance, enabling systematic comparison of different coordination topologies. Our experiments demonstrate that MICA consistently improves task success, reliability, and responsiveness over baseline structures, while remaining deployable on practical offline hardware. Together, these contributions highlight MICA as a step toward deployable, privacy-preserving multi-agent assistants for dynamic factory environments. The source code will be made publicly available at \url{https://github.com/Kratos-Wen/MICA}.
\end{abstract}

%% file: tex/Introduction.tex
\section{Introduction}
\label{sec:intro}

Modern manufacturing increasingly operates under rapid line reconfiguration, product variants, and strict safety and compliance requirements. Assembly procedures are long-horizon and interdependent, with tool–part constraints and exception handling that challenge non-expert and rotating workers; mistakes incur time, quality, and safety costs \cite{capponi2024assembly}. At the same time, privacy and connectivity constraints often preclude cloud offloading, and confidentiality limits the collection of large annotated datasets. 
Although vision-based assistance improves stepwise guidance in realistic settings~\cite{daling2024effects}, reliable on-device deployment under limited data remains difficult.

Large language models have strong general reasoning ability \cite{gu2024survey,yao2024survey}, and multi-agent formulations promise structured problem solving \cite{wu2024autogen,qian2023chatdev,hong2024metagpt,du2023improving,yao2023tree,zhou2022least}. Existing multi-agent evaluations are largely text-centric or simulated, with limited grounding in sensed factory state or speech interaction; coordination reliability degrades under partial or asynchronous observations, conflicting with cycle-time and safety requirements on the shop floor. 
This gap motivates perception-grounded and budget-aware multi-agent assistance. To reduce data and privacy barriers, recent work shows that a small capture of part photos or short multi-view videos, together with manuals, can bootstrap an image and text knowledge base for local, privacy-preserving assistance~\cite{wen2025snap}.

We present \textbf{MICA} (\underline{M}ulti-Agent \underline{I}ndustrial \underline{C}oordination \underline{A}ssistant), a perception-grounded and speech-interactive industrial assistant that runs entirely on edge hardware. MICA couples egocentric vision with multi-agent language reasoning to deliver real-time assembly, troubleshooting, part queries, and maintenance support. The system comprises three integrated modules: \circled{1} Depth-guided Object Context Extraction for stable, view-aligned part context; \circled{2} Adaptive Assembly Step Recognition that blends a state-graph expert with an image-retrieval expert; and \circled{3} \textit{MICA-core}, a modular reasoning layer that routes queries to role-specialized agents under safety auditing. Built on a lightweight image–text knowledge base derived from assembly manuals and a small set of component captures, our system avoids large-scale annotation while remaining adaptable to new assembly procedures.
\begin{figure}[t!]
    \centering
    \framebox{
    \includegraphics[width=0.9\linewidth]{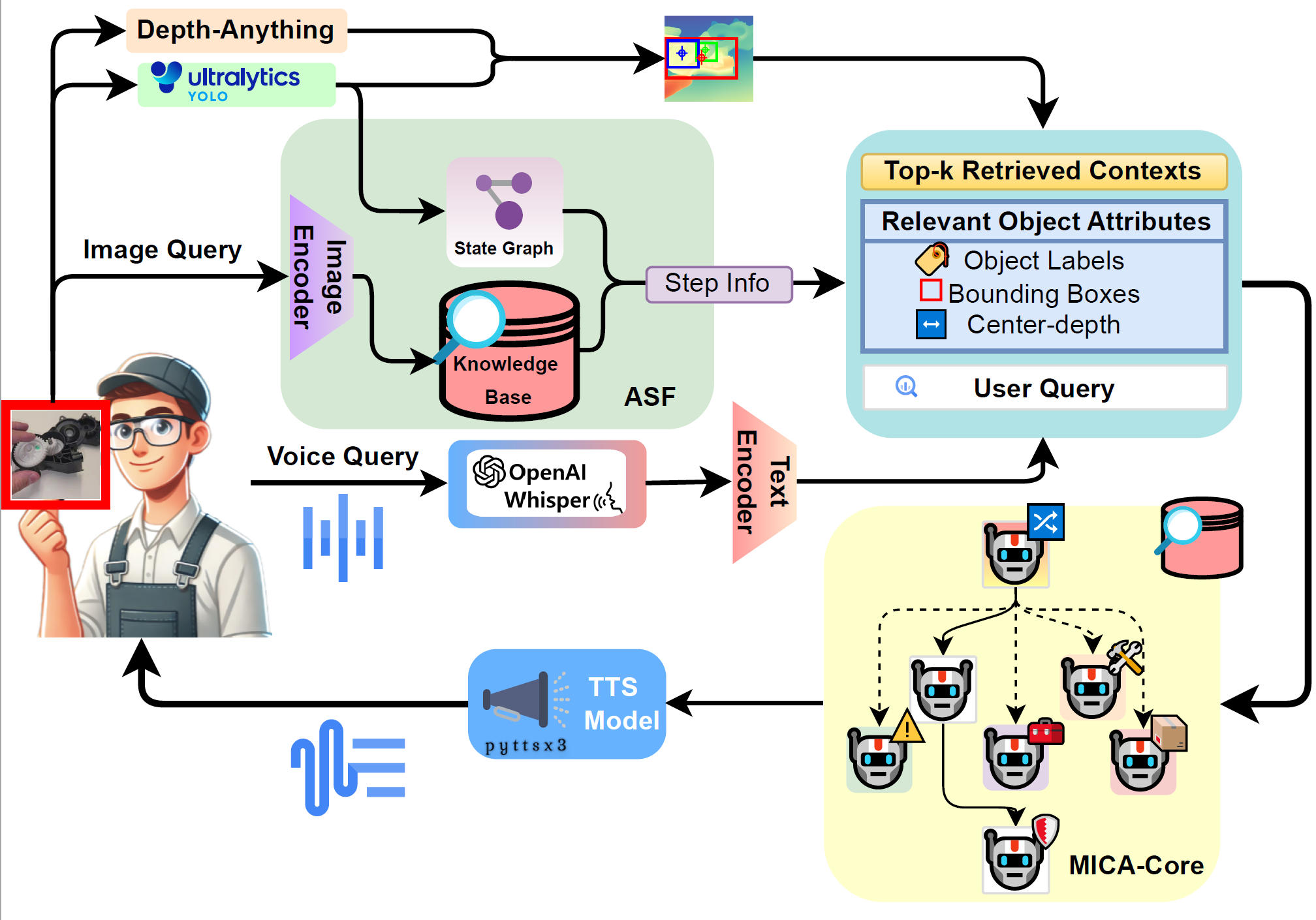}}
    \caption{Overview of the proposed MICA system. Egocentric vision and speech queries are processed into structured object contexts via YOLO-based detection and depth estimation. These contexts, together with state-graph priors and knowledge base information, support Adaptive Step Fusion (ASF) for robust step recognition. The MICA-core then integrates perception and reasoning to deliver safety-audited, speech-based guidance in real time.}
    \label{fig:main_pipeline}
\end{figure}
To enable rigorous comparison under identical tools, prompts, knowledge access, and budgets, we establish a controlled benchmark that instantiates four representative multi-agent topologies. We further introduce two deployment-oriented metrics: Knowledge Base Alignment (KBA) for factual consistency with the curated component knowledge base, and Energy per Successful Answer (E/succ) for energy–utility efficiency in real-time use.

We summarize our contributions as follows:
\begin{itemize}
    \item A fully offline, perception-grounded industrial assistant that unifies egocentric vision, speech I/O, and role-specialized multi-agent reasoning on edge devices.
    \item \textit{Adaptive Step Fusion} (ASF), a lightweight fusion and online adaptation mechanism that integrates rule-based workflow constraints with retrieval-based visual similarity and enables real-time correction through natural-language feedback.
    \item A multi-agent coordination benchmark with standardized protocols and two metrics (KBA and E/succ) tailored to safety-critical industrial assistance.
\end{itemize}

%% file: tex/RelatedWork.tex
\section{Related Work}
\label{sec:related}

\subsection{Real-Time Egocentric Vision in Wearable Systems}
Wearable egocentric vision offers direct access to gaze, hand--object interactions, and short-horizon intent, which are central to shop-floor assistance and safety auditing. Recent surveys and outlooks highlight the growing push toward on-device assistance and privacy-preserving perception~\cite{plizzari2024outlook}. Large-scale benchmarks~\cite{li2025egocross,zhang2025egonight,Damen_2018_epickitchens,Grauman_2022_ego4d}, \eg, EPIC-Kitchens~\cite{Damen_2018_epickitchens} and Ego4D~\cite{Grauman_2022_ego4d}, have catalyzed progress in segmentation, anticipation, and episodic memory, enabling long-horizon reasoning over first-person video. Beyond general benchmarks, task- and modality-focused resources advance the field toward deployment: Nymeria contributes synchronized, multimodal recordings with full-body motion and Aria-based sensors~\cite{ma2024nymeria}; EgoSim provides a multi-view simulator plus real data for body-worn cameras~\cite{hollidt2024egosim}; EgoEnv links first-person video to local environment representations for better state awareness~\cite{nagarajan2023egoenv}. New evaluations target assistance with text and structure, including EgoTextVQA for scene-text-aware video QA and EgoSG for egocentric 3D scene graphs~\cite{zhou2025egotextvqa,zhang2024egosg}. Practical assistive prototypes and wearables illustrate end-user benefits and the value of resource-constrained design~\cite{liu2024objectfinder,zheng2024materobot}. Meanwhile, geometry-aware egocentric scene understanding (e.g., EDINA) addresses tilted viewpoints and dynamic foregrounds common on the factory floor~\cite{do2022egocentric}. Vinci demonstrates an end-to-end egocentric VLM assistant with streaming memory and grounding on portable devices, pointing to real-time, on-device workflows~\cite{vinci}. While recent egocentric resources and prototypes improve on-device perception, many pipelines remain single-model or cloud-assisted, limiting modularity and guaranteed on-device operation in industrial conditions. Our system targets offline, on-device operation by coupling local perception with role-specialized agents and a safety/KB auditor under compute and connectivity constraints.

\subsection{Multi-Agent Large Language Models}
LLMs have moved from single-agent autonomy to multi-agent collaboration, where multiple LLM-based agents communicate, cooperate, or compete to solve tasks beyond a single model’s capacity~\cite{zhang2025webpilot,yu2024fincon,wu2024autogen,qian2023chatdev}. In manufacturing, multi-agent coordination lets distributed machines and software adapt in real time, balance loads, recover from faults, and optimize throughput across heterogeneous equipment—improving flexibility, scalability, and resilience~\cite{wu2024novel,nie2024predictive,wang2024multi,lim2024large,antons2024designing,siatras2024production}. 
Mechanisms that sustain long-horizon interactions include structured roles/memory~\cite{li2023camel,packer2023memgpt} and reasoning curricula that decompose, search, and vote~\cite{yao2023tree,zhou2022least,wang2022self}. Yet most methods remain text-bound without egocentric sensing or actuation, and their coordination reliability degrades under partial/asynchronous observations—conditions at odds with strict cycle-time and safety constraints on the shop floor. Evidence from simulated environments suggests that persistent memory and planning improve long-horizon behavior, while specialization benefits difficult reasoning but may be unnecessary for simple queries~\cite{park2023generative,becker2024multiagentlargelanguagemodels}. To improve coordination, prior work explores open dialogue~\cite{wu2024autogen, qian2023chatdev}, structured workflows ~\cite{hong2024metagpt}, adversarial debate, and learned cooperation modules (COPPER) for cross-verification and refinement~\cite{qian2023chatdev,Yang2025adversarial,du2023improving,bo2024copper}. However, current multi-agent LLM studies are largely evaluated in simulated domains, focusing on communication algorithms rather than real-world perception~\cite{wu2024autogen,hong2024metagpt}. MICA grounds collaboration in sensed state with explicit time/energy budgets and safety auditing, supporting reliable workflows beyond simulated domains.

%% file: tex/Methodology.tex
\section{Methodology}
\label{sec:method}

Our intelligent industrial assistance system, \textbf{MICA} (\underline{M}ulti-Agent \underline{I}ndustrial \underline{C}oordination \underline{A}ssistant), addresses the core challenge of providing accurate, real-time assembly guidance in dynamic factory environments, where visual occlusion, step ambiguity, and safety constraints make robust recognition essential. 
As illustrated in Fig.~\ref{fig:main_pipeline}, the system integrates three tightly coupled modules: (1) \emph{Depth-guided Object Context Extraction}, which focuses on the most relevant components from the worker’s viewpoint; (2) \emph{Adaptive Assembly Step Recognition}, which resolves step ambiguities and adapts online with user feedback; and (3) \emph{Multi-Agent Collaborative Reasoning via MICA-core}, which delivers task-specific guidance under safety auditing. Together, these modules form a pipeline in which perception refines context, step recognition constrains reasoning, and reasoning returns adaptive feedback to the worker.
\subsection{Depth-guided Object Context Extraction}
\label{sec:depth_context_selection}

To ensure reliable perception under dynamic assembly conditions, we adopt YOLOv11~\cite{khanam2024yolov11} as the base detector, trained following~\cite{wen2025snap} on the Gear8 dataset. Each frame produces raw component detections, which are stabilized by aggregating results over a sliding window of $L$ frames. We denote by $\mathbf{b}_i$ the bounding box and by $c_i$ its confidence. Detections with $\mathrm{IoU}(\mathbf{b}_i,\mathbf{b}_j)\!\ge\!\tau_{\mathrm{IoU}}{=}0.5$ are clustered as $\mathcal{C}=\{(\mathbf{b}_i,c_i)\}_{i=1}^{m}$, and fused by confidence-weighted averaging:
\begin{equation}
\hat{\mathbf{b}}=\frac{\sum_{i=1}^{m} c_i\,\mathbf{b}_i}{\sum_{i=1}^{m} c_i}, \qquad
\hat{c}=\tfrac{1}{m}\sum_{i=1}^{m} c_i .
\end{equation}
On this fused result, Depth-Anything~\cite{yang2024depth} estimates pixel-wise depth. The nearest component relative to the camera center in the depth map is taken as the worker’s primary focus, while nearby components within spatial and depth thresholds $(\tau_p,\tau_d)$ are also included to capture peripheral interactions:
\begin{equation}
\mathcal{O}_{\text{rel}} = \{\,o_i \mid \|x_i-x^*\|\le \tau_p,\ |d_i-d^*|\le \tau_d\,\}
\end{equation}
where $(x_i,d_i)$ denote the spatial and depth coordinates of object $o_i$, and $(x^*,d^*)$ correspond to the nearest component. Only this fused, depth-refined context is passed to subsequent modules.

\begin{figure*}[htbp]
    \centering
    \framebox{\includegraphics[width=0.98\textwidth]{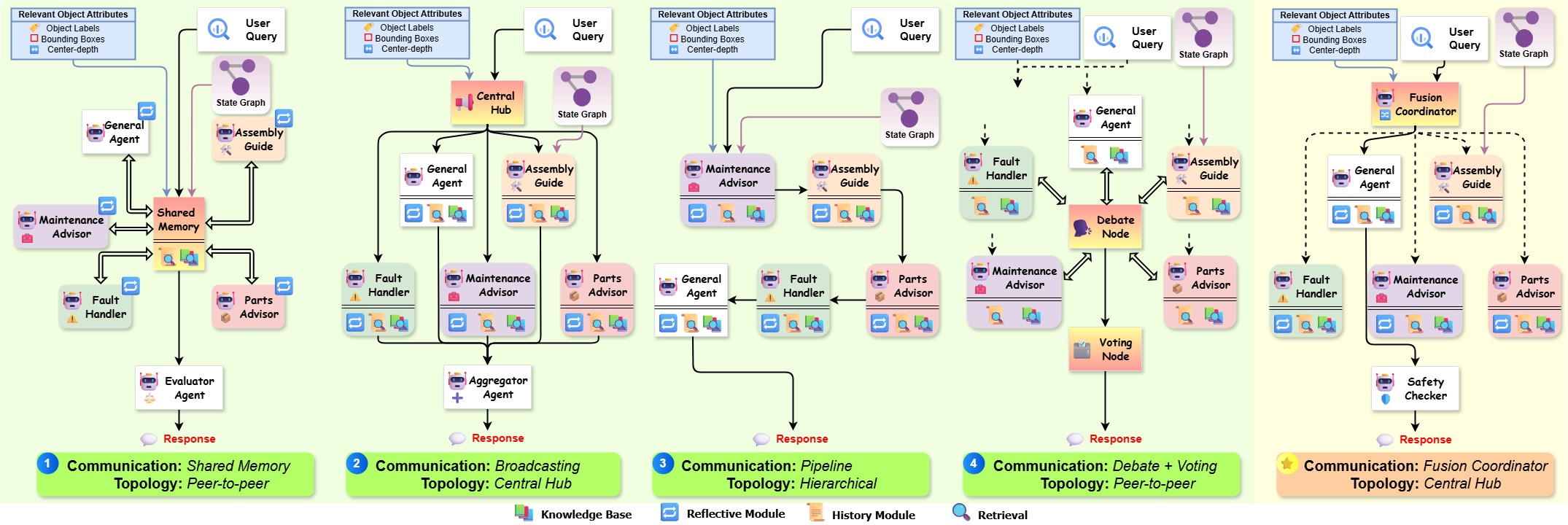}}
    \caption{An overview of the multi-agent LLM baseline architectures for comparison: (1) SharedMemory: decentralized peer-to-peer with a shared memory and evaluator; (2) CentralizedBroadcast: hub-and-spoke publish–subscribe with an aggregator; (3) HierarchicalPipeline: fixed sequential relay across specialists; (4) DebateVoting: peer debate followed by consensus voting.}
    \label{fig:multiagent}
\end{figure*}

\subsection{Adaptive Assembly Step Recognition}
\label{sec:adaptive_step_recognition}

We estimate the assembly step from \emph{streaming} first-person video by integrating two complementary detectors and a lightweight adaptive fusion.
The \emph{state–graph detector} leverages workflow constraints automatically derived from the component knowledge base (KB) and assembly procedure templates to score each candidate step according to required components and their multiplicities, enforcing structural consistency.
The \emph{retrieval detector} compares the current frame against a gallery of reference states in an embedding space to provide a similarity–based estimate.
The two detectors are complementary: the former supplies structure and interpretability, the latter is robust to occlusion and detection noise; our \emph{Adaptive Step Fusion (ASF)} combines them at the class level and adapts online from speech–driven feedback.

\noindent\textbf{\textit{a) State–graph detector.}}
Let $\mathcal{S}=\{S_1,\dots,S_K\}$ be the set of steps. For each step $S_j$, the KB specifies a rule triple $(\mathrm{all\_of}_j,\mathrm{any\_of}_j,\mathrm{forbid}_j)$, where sets list required, alternative, and forbidden components (by KB part IDs).
For brevity, denote $\mathcal{A}_j:=\mathrm{all\_of}_j$, $\mathcal{O}_j:=\mathrm{any\_of}_j$, and $\mathcal{F}_j:=\mathrm{forbid}_j$. Counts $n(k)$ are computed from the depth–refined context $O_{\text{rel}}$ (Sec.~\ref{sec:depth_context_selection}) aggregated over the sliding window; the required multiplicity $r_j(k)$ comes from the KB (default $r_j(k){=}1$ if unspecified).
We score each step by
\begin{equation}
C_s(j)=\alpha\,\mathrm{all}_j+(1-\alpha)\,\mathrm{any}_j-\mathrm{pen}_j,
\label{eq:sg-score}
\end{equation}
where
\begin{align}
\phi_j(k) &:= \min\!\Bigl(1,\frac{n(k)}{\max(1,r_j(k))}\Bigr),\\
\mathrm{all}_j &= \frac{1}{\max(1,|\mathcal{A}_j|)} \sum_{k\in\mathcal{A}_j} \phi_j(k),\\
\mathrm{any}_j &= \mathbb{I}\!\Bigl[\,|\mathcal{O}_j|=0\ \lor\ \exists\,k\in\mathcal{O}_j:\ n(k)\ge r_j(k)\Bigr],\\
\mathrm{pen}_j &= \tfrac{1}{2}\,\mathbb{I}\!\Bigl[\,\exists\,k\in\mathcal{F}_j:\ n(k)>0\Bigr].
\end{align}
Here $\mathbb{I}[\cdot]$ is the indicator, $\alpha\in[0,1]$ (we use $\alpha{=}0.6$).
The detector outputs
\begin{align}
S_s=\arg\max_j C_s(j),\qquad C_s=\max_j C_s(j).
\end{align}

\noindent\textbf{\textit{b) Retrieval detector.}}
Let $f(\cdot)$ be an image encoder~\cite{radford2021learning}, $\{g_j\}$ be per-step references and $q$ denote the current frame.
We score each step by the cosine similarity
\begin{align}
C_r(j) &= \cos\bigl(f(q),\,f(g_j)\bigr)\quad (\text{or top-}k \text{ average}), \label{eq:ret-score}\\
S_r &= \arg\max_j C_r(j),\qquad C_r=\max_j C_r(j). \nonumber
\end{align}

\noindent\textbf{\textit{c) ASF scoring.}}
To fuse the detectors, our \emph{Adaptive Step Fusion (ASF)} maintains per--class expert weights \(W_{j,e}\ge0\), per--class biases \(b_j\), and global gates \(g_e\ge0\) with \(g_s+g_r=1\).
To preserve weak signals from non--winning experts we define
\begin{align}
c_{e,j}=\begin{cases}
C_e, & S_e=S_j,\\
\lambda_e C_e, & \text{otherwise,}
\end{cases}
\quad e\in\{s,r\},
\end{align}
with leak parameters \(\lambda_e\in[0,1)\).
We define the KB coverage as \(\mathrm{cov}_j := \mathrm{all}_j\), i.e., the averaged satisfaction over required components (Sec.~\ref{sec:adaptive_step_recognition}a).
Non--jumping dynamics are encoded by an allowed set \(\mathcal{A}(S_{\mathrm{prev}})\) from the previous fused step. The overall score is
\begin{align}
\label{eq:asf-score}
\mathrm{score}_j
&= b_j + g_s W_{j,s} c_{s,j} + g_r W_{j,r} c_{r,j} \nonumber\\
&\quad + \lambda_{\mathrm{cov}}\,\mathrm{cov}_j
- \lambda_{\mathrm{tr}}\,\mathbb{I}\!\bigl[S_j\notin \mathcal{A}(S_{\mathrm{prev}}))\bigr],
\end{align}
We use nonnegative weights $\lambda_{\mathrm{cov}},\lambda_{\mathrm{tr}}\!\ge\!0$ to balance coverage and transition penalties. The fused step is \(S_f=\arg\max_j \mathrm{score}_j\), with a calibrated confidence obtained by softmax over \(\{\mathrm{score}_j\}\).

\noindent\textbf{\textit{d) ASF online adaptation.}}
User feedback \(y\in\mathcal{S}\) is used to update \((W,b,g)\) without backpropagation.
We define a focal--style impact \(\kappa_e=(1-C_e)^\gamma\) with \(\gamma>0\); confident hits (\(S_e=y, C_e\ge C_{\mathrm{freeze}}\)) are frozen by setting \(\kappa_e=0\).
To reduce collapse into a single class, the effective step size is scaled as
\begin{align}
\eta_{\mathrm{eff}}=\eta\,n_y^{-\rho}\,d,
\end{align}
where \(n_y\) is the number of feedback events on class \(y\), \(\rho\in(0,1]\), and \(d\) depends on the recent fraction of \(y\) in a sliding history window. Let $\hat{\imath}:=\arg\max_{j\neq y}\mathrm{score}_j$ be the highest-scoring non-target class at feedback time. Weights are updated multiplicatively per column within a trust-region bound  \(\tau_{\mathrm{trust}}\):
\begin{align}
W_{y,e} &\leftarrow W_{y,e}\,(1+\delta_{y,e}), \\
W_{\hat\imath,e} &\leftarrow W_{\hat\imath,e}\,(1-\delta_{\hat\imath,e}),
\end{align}
with \(\delta \leftarrow \min\!\big(\eta_{\mathrm{eff}}\kappa_e,\ \tau_{\mathrm{trust}}\big)\).
If both experts err, we correct only the column with lower \(C_e\) to avoid oscillation.
Biases are adjusted conservatively with conservation across classes and clipped to \(|b_j|\le b_{\max}\).
Gates are nudged only when exactly one expert hits and then renormalized to \(g_s+g_r=1\).
After each update, columns \(\{W_{j,e}\}_j\) are clamped and renormalized, and a floor \(W_{j,e}\ge\varepsilon_{\mathrm{floor}}\) avoids starving classes.
All parameters are persisted and warm–started across sessions.
Together, ASF introduces three key innovations: (i) explicit incorporation of workflow compatibility and non-jumping transitions into the fusion score, (ii) class–wise fusion with confidence–aware online updates, and (iii) anti–collapse regularization through history–based scaling and weight floors. These choices provide a lightweight yet effective mechanism for online adaptation in streaming assembly recognition.

\subsection{Multi-Agent Collaborative Reasoning via MICA-core}
\label{sec:mica_core}

To transform raw perceptual signals into actionable guidance, we introduce \emph{MICA-core}, a modular multi-agent reasoning framework built on an instruction-tuned LLM~\cite{qwen2025qwen25technicalreport}. MICA-core receives structured inputs from the preceding modules, namely (i) object contexts from depth-guided detection, and (ii) assembly step hypotheses from ASF. Together with natural-language queries transcribed by Speech-to-Text (STT)~\cite{radford2023robust}, these signals form a unified reasoning context. 

Within MICA-core, a lightweight LLM router dynamically assigns each query to one of five specialized agents: \emph{Assembly Guide}, \emph{Parts Advisor}, \emph{Maintenance Advisor}, \emph{Fault Handler}, and a fallback \emph{General Agent}. Each agent operates under a Retrieval-Augmented Generation (RAG) paradigm, retrieving agent-specific evidence from the structured KB and refining responses through iterative reasoning.

To guarantee reliability in safety-critical assembly contexts, all agent outputs are audited by a dedicated safety checker that enforces rule-based assembly constraints and verifies responses against the KB. This layer enforces domain constraints such as correct tool usage, assembly order, and hazard warnings, thereby preventing unsafe recommendations from reaching the user. The combination of dynamic routing, specialized RAG agents, and explicit safety auditing allows MICA-core to deliver contextually precise, semantically rich, and industrially safe responses.

\subsection{Speech-based Interactive Feedback Loop}
\label{sec:speech_feedback}

The reasoning outputs of MICA-core are embedded into an interactive feedback loop with the worker. Queries are captured via Speech-to-Text (STT)~\cite{radford2023robust}, while system responses and status updates are synthesized through Text-to-Speech (TTS)~\cite{c63}. Crucially, workers can verbally confirm or correct ASF’s step predictions in real time, directly influencing the online adaptation of the fusion module. This human-in-the-loop mechanism improves recognition accuracy while making the adaptation process explicit to the worker.

%% file: tex/Evaluation.tex
\section{Experiments}
\label{sec:experiments}

\subsection{Implementation Details}

Experiments are implemented in PyTorch~2.6.0 with CUDA~12.4. The YOLOv11-L detector~\cite{khanam2024yolov11} is fine-tuned on Gear8 following~\cite{wen2025snap}. Multi-frame fusion uses $\tau_{\mathrm{IoU}}=0.5$, confidence threshold $0.4$, at least $m=3$ detections, and persistence over $T=5$ consecutive frames.
Depth estimation is performed with Depth-Anything-V2-Large~\cite{yang2024depth}, using spatial and depth thresholds $(\tau_p,\tau_d)$ for context refinement. 
In ASF, we set $\alpha=0.6$ (rule balance), focal factor $\gamma=2$, base step size $\eta=0.1$ and history scaling $\rho=0.5$. 
Regularization includes trust-region bound $\tau_{\mathrm{trust}}=0.2$, bias bound $b_{\max}=1.0$, and weight floor $\varepsilon_{\mathrm{floor}}=10^{-3}$. Semantic retrieval uses SentenceTransformer (all-MiniLM-L6-v2)~\cite{reimers2019sentence} with FAISS~\cite{douze2024faiss}, and multi-agent reasoning uses Qwen2.5-7B-Instruct~\cite{qwen2025qwen25technicalreport}. Speech recognition uses Whisper-small~\cite{radford2023robust} (16\,kHz, 8\,s windows), and TTS uses pyttsx3~\cite{c63} (180\,wpm, 22.05\,kHz).

\subsection{Multi-Agent Coordination Benchmark}
\label{sec:baselines}

To systematically study coordination under identical tools, prompts, KB access, and backbone LLM, we establish a controlled benchmark comprising four representative interaction topologies (Fig.~\ref{fig:multiagent}). Each topology is instantiated as an engineering counterpart of a well-studied paradigm, providing a standardized protocol for fair comparison.

\noindent\textbf{\textit{SharedMemory}} (Fig.~\ref{fig:multiagent}\,(1)).  
Peer agents read and write a shared blackboard context, submit independent proposals, and a separate evaluator selects the final answer~\cite{li2023camel,packer2023memgpt}.

\noindent\textbf{\textit{CentralizedBroadcast}} (Fig.~\ref{fig:multiagent}\,(2)).  
A central hub broadcasts the task state to all agents, collects parallel responses, and aggregates them into a single output~\cite{wu2024autogen,qian2023chatdev}.

\noindent \textbf{\textit{HierarchicalPipeline}} (Fig.~\ref{fig:multiagent}\,(3)).  
Agents are arranged in a fixed relay, where each stage refines the previous output before passing it to the next~\cite{zhou2022least,yao2023tree}.

\noindent\textbf{\textit{DebateVoting}} (Fig.~\ref{fig:multiagent}\,(4)).  
Agents independently draft responses, critique one another, and then vote to select a consensus output~\cite{du2023improving,wang2022self}.

We evaluate all topologies on five task categories (\textit{General}, \textit{Assembly}, \textit{Part Attributes}, \textit{Maintenance}, and \textit{Fault Handling}) under identical compute budgets and the same knowledge grounding, which enables a controlled assessment of coordination efficacy.

\begin{table}[t]
\centering
\caption{Per-step performance of ASF before and after online adaptation (10 updates per step). Best results in each column are bold; in case of ties, all best entries are bold.}
\label{tab:asf_before_after}
\resizebox{\columnwidth}{!}{
\begin{tabular}{lcccccccccc}
\toprule
\multirow{2}{*}{\textbf{Step}} &
\multicolumn{2}{c}{\textbf{Acc~(\%)~$\uparrow$}} & \multicolumn{2}{c}{\textbf{Prec~(\%)~$\uparrow$}} &
\multicolumn{2}{c}{\textbf{Rec~(\%)~$\uparrow$}} & \multicolumn{2}{c}{\textbf{F1~(\%)~$\uparrow$}} & \multicolumn{2}{c}{\textbf{ECE~$\downarrow$}} \\
\cmidrule(lr){2-3}\cmidrule(lr){4-5}\cmidrule(lr){6-7}\cmidrule(lr){8-9}\cmidrule(lr){10-11}
 & \textbf{Baseline} & \textbf{w/ ASF} & \textbf{Baseline} & \textbf{w/ ASF} & \textbf{Baseline} & \textbf{w/ ASF} & \textbf{Baseline} & \textbf{w/ ASF} & \textbf{Baseline} & \textbf{w/ ASF}\\
\midrule
 S1 & \textbf{97.63} & 92.71 & 70.17 & \textbf{85.37} & \textbf{97.63} & 96.84 & 81.65 & \textbf{90.74} & 0.49 & \textbf{0.38}  \\
 S2 & 81.82 & \textbf{81.88} & \textbf{91.45} & 90.68 & \textbf{77.54} & \textbf{77.54} & \textbf{83.92} & 83.59 & \textbf{0.50} & \textbf{0.50} \\
 S3 & 88.98 & \textbf{97.08} & \textbf{97.41} & 95.73 & \textbf{88.98} & 88.19 & \textbf{93.00} & 91.80 & 0.55 & \textbf{0.54} \\
 S4 &  0.00 & \textbf{95.34} &  0.00 & \textbf{91.46} &  0.00 & \textbf{89.29} &  0.00 & \textbf{90.36} & 0.52 & \textbf{0.43} \\
\midrule
\end{tabular}}
\end{table}

\begin{table*}[htbp]
    \centering
    \caption{Benchmark results for \textit{MICA-core} and four coordination topologies across five categories (General, Assembly-related, Part Attribute, Maintenance-related, Fault Handling). Best results in each column are bold; in case of ties, all best entries are bold. KBA is not computed for the \emph{General Question} because it lacks a structured KB alignment target.}
    \label{tab:multi_agent_evaluation}
    \resizebox{\textwidth}{!}{
    \begin{tabular}{lcccccccccccc}
        \toprule
        \multirow{2}{*}{\textbf{Topology}} 
        & \multicolumn{4}{c}{\textbf{Automatic Evaluation Metrics}} 
        & \multicolumn{5}{c}{\textbf{GPT-based Evaluation Metrics}}
        & \multirow{2}{*}{\textbf{AL (s)~$\downarrow$}} & \multirow{2}{*}{\textbf{E/succ (kJ)~$\downarrow$}} \\
        \cmidrule(lr){2-5} \cmidrule(lr){6-10}
        & \textbf{TS (\%)~$\uparrow$} & \textbf{BL~$\uparrow$} & \textbf{RG~$\uparrow$} & \textbf{KBA (\%)~$\uparrow$}
        & \textbf{Acc~$\uparrow$} & \textbf{Rel~$\uparrow$} & \textbf{Con~$\uparrow$} & \textbf{Help~$\uparrow$} & \textbf{Safe~$\uparrow$} &  \\
        \midrule
        & \multicolumn{9}{c}{\textbf{General Question}} &  \\
        SharedMemory & 31.25 & 0.52 & 0.50 & N/A & 3.25 & 3.19 & \textbf{4.88} & 2.91 & \textbf{5.00} & 2.79 & 2.03\\
        CentralizedBroadcast & 12.50 & 0.46 & 0.44 & N/A & 2.22 & 3.25 & 4.47 & 2.66 & \textbf{5.00} & 2.87 & 5.13\\
        HierarchicalPipeline & 23.13 & 0.48 & 0.50 & N/A & 3.12 & 3.25 & 3.78 & 3.22 & \textbf{5.00} & 2.92 & 2.33 \\
        DebateVoting & 59.38 & 0.64 & 0.65 & N/A & 3.31 & 4.22 & 4.56 & 3.75 & \textbf{5.00} & 5.08 & 2.66 \\
        \rowcolor{gray!20} MICA-core (ours) & \textbf{90.63} & \textbf{0.76} & \textbf{0.77} & N/A & \textbf{4.19} & \textbf{4.41} & 4.59 & \textbf{4.25} & \textbf{5.00} & \textbf{0.58} & \textbf{0.71}\\
        & \multicolumn{9}{c}{\textbf{Assembly-related Question}} & \\
        SharedMemory & 15.63 & 0.07 & 0.09 & 9.10 & 1.94 & 2.16 & 2.91 & 1.84 & \textbf{5.00} & 3.77 & 5.07\\
        CentralizedBroadcast & \textbf{46.88} & \textbf{0.24} & \textbf{0.36} & 19.10 & \textbf{3.00} & \textbf{2.94} & \textbf{4.94} & 2.75 & \textbf{5.00} & 4.19 & 2.31 \\
        HierarchicalPipeline & 22.88 & 0.13 & 0.24 & 10.24 & 2.03 & 2.22 & 3.31 & 2.19 & \textbf{5.00} & 3.88 & 4.68 \\
        DebateVoting & 37.50 & 0.07 & 0.18 & 5.43 & 2.12 & 2.31 & 3.88 & 2.28 & \textbf{5.00} & 7.51 & 4.30 \\
        \rowcolor{gray!20} MICA-core (ours) & 43.75 & 0.19 & 0.28 & \textbf{21.18} & 2.91 & 2.88 & \textbf{4.94} & \textbf{2.78} & \textbf{5.00} & \textbf{0.74} & \textbf{1.61} \\
        & \multicolumn{9}{c}{\textbf{Part Attribute Question}} & \\
        SharedMemory & 78.13 & 0.12 & 0.06 & 9.32 & 2.16 & 2.53 & 3.84 & 2.47 & \textbf{5.00} & 4.06 & 1.20 \\
        CentralizedBroadcast & 71.88 & 0.50 & 0.58 & \textbf{36.98} & 3.91 & 4.09 & \textbf{4.91} & 4.03 & \textbf{5.00} & 3.95 & 1.84\\
        HierarchicalPipeline & 93.75 & 0.20 & 0.27 & 33.57 & 3.97 & 4.00 & 4.47 & 3.91 & \textbf{5.00} & 3.95 & 1.27\\
        DebateVoting & \textbf{96.88} & \textbf{0.57} & \textbf{0.62} & 30.68 & 4.12 & 4.03 & 4.88 & \textbf{4.09} & \textbf{5.00} & 7.92 & 1.40\\
        \rowcolor{gray!20} MICA-core (ours) & \textbf{96.88} & 0.38 & 0.45 & 36.68 & \textbf{4.22} & \textbf{4.12} & 4.59 & \textbf{4.09} & \textbf{5.00} & \textbf{0.77} & \textbf{0.73}\\
        & \multicolumn{9}{c}{\textbf{Maintenance-related Question}} & \\
        SharedMemory & \textbf{37.50} & 0.11 & 0.06 & \textbf{10.53} & \textbf{2.31} & \textbf{2.22} & 3.06 & \textbf{1.97} & \textbf{5.00} & 3.28 & 5.36 \\
        CentralizedBroadcast & 25.00 & \textbf{0.24} & \textbf{0.36} & 5.10 & 2.03 & 2.12 & 2.94 & 1.72 & \textbf{5.00} & 3.31 & \textbf{3.64} \\
        HierarchicalPipeline & 6.25 & 0.06 & 0.09 & 3.04 & 1.03 & 1.19 & \textbf{3.62} & 1.12 & \textbf{5.00} & 3.11 & 10.29\\
        DebateVoting& 12.50 & 0.04 & 0.06 & 9.09 & 1.16 & 1.22 & 3.56 & 1.25 & \textbf{5.00} & 6.85 & 46.11 \\
        \rowcolor{gray!20} MICA-core (ours) & 21.88 & 0.05 & 0.08 & 5.13 & 1.22 & 1.38 & 2.66 & 1.47 & \textbf{5.00} & \textbf{0.78} & 5.67\\
        & \multicolumn{9}{c}{\textbf{Fault Handling Question}} & \\
        SharedMemory & 56.25 & 0.23 & 0.35 & 5.45 & 2.91 & 3.00 & 3.84 & 3.19 & \textbf{5.00} & 3.76 & 2.51\\
        CentralizedBroadcast & 46.88 & \textbf{0.24} & \textbf{0.36} & 5.98 & 3.03 & 3.09 & 3.56 & 3.03 & \textbf{5.00} & 3.59 & 1.78\\
        HierarchicalPipeline & \textbf{62.50} & 0.22 & \textbf{0.36} & 5.84 & 3.19 & 3.16 & 3.84 & 3.06 & \textbf{5.00} & 3.84 & 2.04\\
        DebateVoting & 50.00 & 0.12 & 0.24 & 5.13 & 3.03 & 3.12 & 2.91 & 3.06 & \textbf{5.00} & 7.48 & 2.89\\
        \rowcolor{gray!20} MICA-core (ours) & \textbf{62.50} & 0.13 & 0.25 & \textbf{13.49} & \textbf{3.31} & \textbf{3.84} & \textbf{4.34} & \textbf{3.69} & \textbf{5.00} & \textbf{0.68} & \textbf{1.53}\\
        \midrule
        & \multicolumn{9}{c}{\textbf{Overall Average}} & \\
        \midrule
        SharedMemory & 43.75 & 0.21 & 0.21 & 8.60 & 2.51 & 2.62 & 3.71 & 2.48 & \textbf{5.00} & 3.53 & 3.23 \\
        CentralizedBroadcast 
        & 40.63 & \textbf{0.34} & \textbf{0.42} & 16.79 & 2.84 & 3.10 & 4.16 & 2.84 & \textbf{5.00} & 3.58 & 2.94 \\
        HierarchicalPipeline 
        & 41.70 & 0.22 & 0.29 & 13.17 & 2.67 & 2.76 & 3.80 & 2.70 & \textbf{5.00} & 3.54 & 4.12 \\
        DebateVoting 
        & 51.25 & 0.29 & 0.35 & 12.58 & 2.75 & 2.98 & 3.96 & 2.89 & \textbf{5.00} & 6.97 & 11.47 \\
        \rowcolor{gray!20} MICA-core (ours) 
        & \textbf{63.13} & 0.30 & 0.37 & \textbf{19.12} & \textbf{3.17} & \textbf{3.33} & \textbf{4.22} & \textbf{3.26} & \textbf{5.00} & \textbf{0.71} & \textbf{2.05} \\
        \bottomrule
    \end{tabular}}
\end{table*}

\subsection{Evaluation Metrics and Setup}
\label{sec:eval}

We evaluate (i) the effect of online Adaptive Step Fusion (ASF, Sec.~\ref{sec:adaptive_step_recognition}) and (ii) the comparative performance of the multi-agent topologies (Sec.~\ref{sec:baselines}).

\noindent\textbf{\textit{a) ASF evaluation.}}
We report pre/post-adaptation performance on step prediction $S_f$ using accuracy (Acc), precision (Prec), recall (Rec), F1-score (F1), and Expected Calibration Error (ECE)~\cite{guo2017calibration}, which measures the alignment between predicted confidence and empirical correctness.

\noindent\textbf{\textit{b) Benchmark protocol and metrics.}}
We evaluate the four benchmark topologies using three families of metrics (Tab.~\ref{tab:multi_agent_evaluation}):

\noindent\emph{(i) Automatic evaluation metrics.}
We use three automatic metrics: (a) task success (TS, \%), a binary indicator of whether an answer satisfies the task-specific success criterion defined by deterministic KB-derived rules; (b) BLEU (BL)~\cite{papineni2002bleu} and ROUGE-L (RG)~\cite{lin2004rouge}, which measure lexical and subsequence overlap with reference responses; and (c) \emph{Knowledge Base Alignment} (KBA, \%), a benchmark-specific metric for factual consistency with the curated component KB.
Given an answer $a$, we extract canonical KB phrases appearing in $a$ and compute the coverage of KB attribute categories referenced by these phrases. Let $P(a)$ denote phrase precision and $R(a)$ the fraction of covered attribute categories. The final KBA score is defined as the harmonic mean
\[
\mathrm{KBA}(a)=\frac{2P(a)R(a)}{P(a)+R(a)} .
\]

\noindent\emph{(ii) GPT-based evaluation.}
Following recent LLM evaluation practice~\cite{zhou2023lima,wang2023chatgpt}, we use GPT-4o~\cite{hurst2024gpt} as a judge to score factual accuracy (Acc), relevance (Rel), consistency (Con), helpfulness (Help), and safety (Safe).

\noindent\emph{(iii) Resource-oriented metrics.}
We report end-to-end Average Latency (AL, s), measured from the availability of the ASF output to completion of the assistant's response, and Energy per Successful Answer (E/succ, kJ), computed from GPU power measurements collected via NVIDIA NVML after subtracting the idle baseline.

\noindent\textbf{\textit{c) Experimental setup.}}  

\noindent\emph{(i) ASF adaptation.} We consider four assembly steps with annotated ground truth. Pre-adaptation uses initial ASF parameters; post-adaptation is measured after ten updates per step. The ten-update budget balances operator effort and adaptation efficacy in industrial workflows.

\noindent\emph{(ii) Benchmark evaluation.} To isolate coordination effects, we use fixed video segments and ground-truth labels as inputs, thereby removing perception noise from the comparison. The Gear8 dataset~\cite{wen2025snap} contains eight components; for each component and category, we formulate four queries, yielding 32 queries per category (160 in total across five categories: general, assembly, attributes, maintenance, and fault handling). All topologies are evaluated under identical budgets and knowledge grounding. Unless otherwise noted, all LLM calls use deterministic decoding with fixed prompts and a frozen KB snapshot, without self-consistency sampling or retries.

\begin{figure*}[ht]
  \centering
  \framebox{\includegraphics[width=\textwidth, trim={0.01\textwidth} 0 0 0, clip]{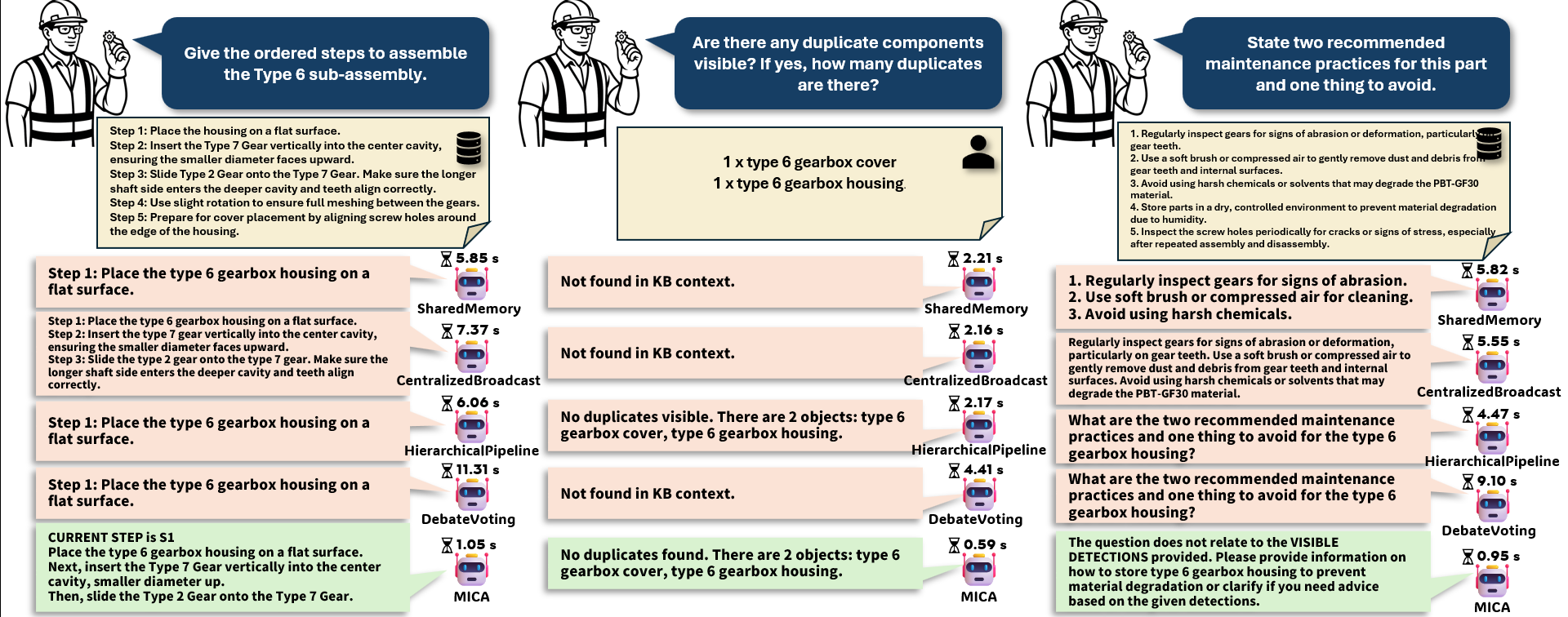}}
  \caption{Qualitative comparison of four representative multi-agent topologies (\textit{SharedMemory}, \textit{CentralizedBroadcast}, \textit{HierarchicalPipeline}, \textit{DebateVoting}) against \textit{MICA} on three representative queries.}
  \label{fig:qualitative}
\end{figure*}

\subsection{Quantitative Results}
\label{sec:expriments:quantitative}
We report the impact of ASF adaptation on step recognition and present a controlled comparison of coordination topologies across five categories, evaluated by automatic, GPT-based, and efficiency metrics.

\noindent\textbf{\textit{a) ASF adaptation.}}  
Tab.~\ref{tab:asf_before_after} demonstrates that online ASF substantially improves robustness with minimal feedback. Ten lightweight corrections per step reduce calibration error (ECE) across all steps and correct systematic late-stage failures. In particular, $S_4$ improves from $0\%$ to $95.34\%$ accuracy and achieves $+90.36$ F1, showing that feedback-driven reweighting is most effective when the state graph and retrieval detector diverge. Mid-sequence steps ($S_3$) also benefit with $+8.1$ accuracy and reduced ECE ($0.55\rightarrow0.54$). By contrast, $S_1$ saturates quickly: accuracy drops marginally ($97.63\%\rightarrow92.71\%$) while precision rises by $+15.2$, reflecting a precision–recall rebalancing. These dynamics confirm ASF’s practicality: a small, fixed supervision budget converts an initially brittle fusion into a calibrated, generalizable predictor without requiring prolonged operator involvement.

\noindent\textbf{\textit{b) Benchmarking coordination topologies.}}  
Tab.~\ref{tab:multi_agent_evaluation} benchmarks MICA-core against four representative coordination structures under controlled conditions. On average, MICA achieves the highest task success (TS $63.13\%$) and the strongest knowledge base alignment (KBA $19.12\%$), while maintaining the lowest latency ($0.71$\,s) and the lowest energy per successful answer ($2.05$\,kJ). This profile indicates that \textit{MICA} balances factual faithfulness, responsiveness, and efficiency, whereas baselines tend to sacrifice at least one of these dimensions.

\noindent\emph{(i) Category-specific behaviors.}  
Performance varies by query type. \textit{SharedMemory} is strongest on maintenance (TS $37.50\%$), where co-occurrence heuristics align with routine safety checks, but it exhibits high latency due to evaluator overhead and generalizes poorly beyond this category. \textit{CentralizedBroadcast} peaks on assembly (TS $46.88\%$), benefiting from synchronized access to step context, at the cost of higher energy consumption caused by parallel yet redundant agent activations. \textit{DebateVoting} excels on part attributes (TS $96.88\%$, BL $0.57$, RG $0.62$), where surface lexical correctness dominates and peer critique can sharpen phrasing; however, it degrades on assembly and maintenance, as repeated critique on partially incorrect premises amplifies noise and increases latency and energy. \textit{HierarchicalPipeline} delivers coherent but brittle outputs: once an upstream error occurs, downstream agents have no mechanism to correct it, which explains its stable yet moderate scores. MICA leads on general (TS $90.63\%$) and fault handling (TS $62.50\%$) through KB grounding and adaptive routing; its conservative router occasionally under-recalls in maintenance, which accounts for the weaker relative score in that category.

\noindent\emph{(ii) Error modes and router sensitivity.}  
Failure cases reveal diagnostic patterns. Ambiguous phrasing and domain synonyms weaken intent signals and lead to conservative routing to a KB-grounded agent. The answer remains factual but may be incomplete relative to the success criterion, reducing TS. \textit{SharedMemory} benefits from accumulated cross-agent co-occurrence, while \textit{CentralizedBroadcast} mitigates misrouting by exposing the same context to all agents, although both incur higher latency or energy.

\noindent\emph{(iii) BLEU/ROUGE versus grounded quality.} \textit{CentralizedBroadcast} attains higher BLEU/ROUGE (0.34/0.42) than MICA (0.30/0.37) yet underperforms in KBA (16.79\% versus 19.12\%). This reflects a tendency to produce longer, templated responses that overlap lexically with references but deviate from KB facts. MICA enforces canonical terminology and safety auditing, yielding concise, action-oriented outputs with lower surface overlap yet stronger factual alignment. GPT-based judgments confirm that lexical overlap is an unreliable proxy for procedural quality in safety-critical tasks, motivating the use of KBA in this benchmark.

\noindent\emph{(iv) Efficiency and utility.} 
Resource measurements reveal clear trade-offs. \textit{DebateVoting} and \textit{CentralizedBroadcast} incur high latency (6.97\,s and 3.58\,s) and energy (11.47\,kJ and 2.94\,kJ), consistent with redundant agent activations. \textit{SharedMemory} also suffers high latency (3.53\,s) due to evaluator overhead. MICA’s sparse activation yields approximately 0.71\,s responsiveness and the lowest energy cost (2.05\,kJ). These results expose a three-way frontier among grounded quality, coordination accuracy, and efficiency; \textit{MICA} occupies the region most suitable for deployment.

Overall, the benchmark shows that while individual baselines exhibit narrow advantages, MICA uniquely balances factual alignment, efficiency, and adaptivity for real-world assistance.

\subsection{Qualitative Results}
To complement the quantitative benchmark, we present targeted case studies that reveal how perception grounding, router-based specialization, and ASF shape system behavior across distinct query types. 
Fig.~\ref{fig:qualitative} compares \textit{SharedMemory}, \textit{CentralizedBroadcast}, \textit{HierarchicalPipeline}, \textit{DebateVoting}, and \textit{MICA} on three queries.

\noindent\textbf{\textit{a) Assembly-related.}} The KB contains a canonical sequence, yet \textit{MICA} does not copy it. The router sends the query to the Assembly Guide, which conditions on detected components and rewrites the steps into a concise, user-oriented list rather than a raw KB block. \textit{SharedMemory} and \textit{CentralizedBroadcast} often yield truncated or fragmented sequences due to evaluator selection and hub aggregation; \textit{HierarchicalPipeline} propagates early omissions; \textit{DebateVoting} increases delay without gains. These outcomes follow the coordination mechanics: single-agent routing in \textit{MICA}, shared evaluator, hub aggregation, fixed relays, and peer debate with voting.

\noindent\textbf{\textit{b) General.}} The duplicate-check has no KB entry and must rely on perception. \textit{MICA} correctly reports two distinct objects with no duplicates by routing to a generalist agent that answers from detections, avoiding reliance on retrieval. Baselines fail with “not found in KB” or misread detections because their decision paths prioritize retrieval and cross-agent aggregation over perception-grounded routing.

\noindent\textbf{\textit{c) Maintenance-related.}} A diagnostic failure occurs when \textit{MICA} misroutes to a detection-focused agent, producing a factual but intent-mismatched answer. The safety checker still audits outputs and prevents unsafe advice. \textit{SharedMemory} and \textit{CentralizedBroadcast} succeed by exposing the same KB content to multiple specialists and selecting or merging a maintenance response, at the cost of higher latency. This shows the trade-off: sparse routing in \textit{MICA} yields efficiency and interpretability, yet intent ambiguity can reduce task success if dispatch is incorrect.

Overall, these cases highlight the synergy between ASF-driven procedural grounding and router-based specialization in \textit{MICA}: with KB support, steps are reformulated for clearer execution; without KB, perception-grounded routing yields accurate answers; and when routing errs, failures remain attributable and auditable.

%% file: tex/Discussion.tex
\section{Conclusion}
\label{sec:conclusion}

We presented MICA, a multi-agent industrial coordination assistant that unifies perception-grounded reasoning, adaptive step understanding, and speech-based interaction for real-time factory support. Our contributions include Adaptive Step Fusion (ASF), which enables continual step-level adaptation through expert blending and speech feedback, and a benchmark with tailored evaluation metrics for systematic comparison of multi-agent coordination strategies. Experiments show that MICA consistently improves task success, reliability, and responsiveness over representative baselines while remaining practical for offline deployment on resource-constrained hardware. Beyond these gains, MICA suggests a pathway toward deployable, privacy-preserving industrial assistants capable of adapting to dynamic workflows. Future work will extend user studies, improve robustness under perception noise and industrial acoustic conditions, and explore deployment on embedded edge platforms.